%% file: PaperForReview.tex
\pdfoutput=1
\documentclass[10pt,twocolumn,letterpaper]{article}

\usepackage{amssymb}
\usepackage[pagenumbers]{cvpr} 
\usepackage{lineno}
\usepackage{wasysym}
\usepackage{marvosym}

\usepackage{graphicx}
\usepackage{amsmath}
\usepackage{amssymb}

\usepackage{tabularx}
\usepackage{adjustbox}
\usepackage{times}
\usepackage{booktabs}
\usepackage{microtype}
\usepackage{epsfig}
\usepackage[table,xcdraw]{xcolor}
\usepackage{caption}
\usepackage{float}
\usepackage{placeins}
\usepackage{color, colortbl}
\usepackage{stfloats}
\usepackage{tcolorbox}
\usepackage{enumitem}
\usepackage{tabularx}
\usepackage{xstring}
\usepackage{xspace}
\usepackage{url}
\usepackage{subcaption}
\usepackage{multirow}
\usepackage{multicol}
\usepackage{xcolor}
\usepackage[hang,flushmargin]{footmisc}
\usepackage{makecell}

\usepackage[normalem]{ulem}
\useunder{\uline}{\ul}{}
\usepackage{booktabs}
\usepackage[switch]{lineno}

\definecolor{cvprblue}{rgb}{0.21,0.49,0.74}
\usepackage[pagebackref,breaklinks,colorlinks,citecolor=cvprblue]{hyperref}

\usepackage[capitalize]{cleveref}
\crefname{section}{Sec.}{Secs.}
\Crefname{section}{Section}{Sections}
\Crefname{table}{Table}{Tables}
\crefname{table}{Tab.}{Tabs.}

\def\paperID{14959} 
\def\confName{CVPR}
\def\confYear{2025}

\newcommand{\best}[1]{$\mathbf{#1}$}
\newcommand{\secondbest}[1]{$\underline{#1}$}

\title{SPAgent: Adaptive Task Decomposition and Model Selection for General Video Generation and Editing}

\author{
Rong-Cheng Tu$^{1*}$ \quad
Wenhao Sun$^{1*}$ \quad 
Zhao Jin$^{1}$\thanks{\quad Equal contributions} \quad 
Jingyi Liao$^{1}$ \\
Jiaxing Huang$^{1}$ \quad 
Dacheng Tao$^{1}$\thanks{\quad Corresponding author}\\
$^{1}$~Nanyang Technological University, Singapore\\
    \texttt{\{rongcheng.tu,wenhao006\}@ntu.edu.sg}, \texttt{jz476051@gmail.com}, \\ 
    \texttt{\{jingyi012,jiaxing.huang,dacheng.tao\}@ntu.edu.sg}
}

\makeatletter
\AtBeginDocument{
\let\@oldmaketitle\@maketitle
\renewcommand{\@maketitle}{\@oldmaketitle
  \includegraphics[width=\linewidth]{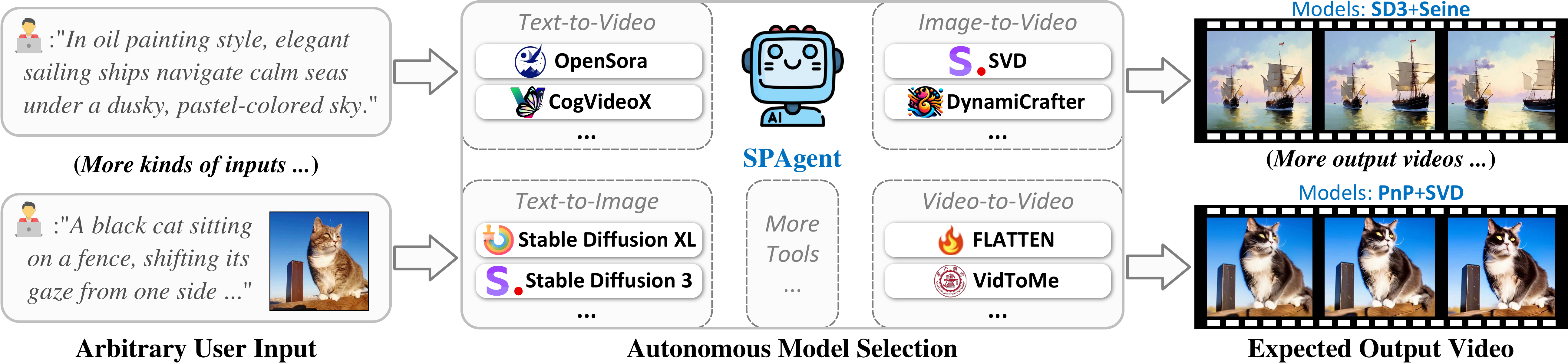}
  \captionof{figure}{Illustration of our Semantic Planning Agent (SPAgent) for general video generation and editing. 
    The SPAgent system is versatile and can adaptively handle a variety of video generation and editing tasks.
  }
  \label{fig:teaser}  
  \vspace{17pt}
 }
}
\makeatother

\begin{document}
\maketitle


\begin{abstract}
While open-source video generation and editing models have made significant progress, individual models are typically limited to specific tasks, failing to meet the diverse needs of users. Effectively coordinating these models can unlock a wide range of video generation and editing capabilities. However, manual coordination is complex and time-consuming, requiring users to deeply understand task requirements and possess comprehensive knowledge of each model's performance, applicability, and limitations, thereby increasing the barrier to entry.
To address these challenges, we propose a novel video generation and editing system powered by our Semantic Planning Agent (SPAgent). SPAgent bridges the gap between diverse user intents and the effective utilization of existing generative models, enhancing the adaptability, efficiency, and overall quality of video generation and editing. Specifically, the SPAgent assembles a tool library integrating state-of-the-art open-source image and video generation and editing models as tools.
After fine-tuning on our manually annotated dataset, SPAgent can automatically coordinate the tools for video generation and editing, through our novelly designed three-step framework: (1) decoupled intent recognition, (2) principle-guided route planning, and (3) capability-based execution model selection. 
Additionally, we enhance the SPAgent's video quality evaluation capability, enabling it to autonomously assess and incorporate new video generation and editing models into its tool library without human intervention.  Experimental results demonstrate that the SPAgent effectively coordinates models to generate or edit videos, highlighting its versatility and adaptability across various video tasks.
\end{abstract}

\input{sections/1introduction}
\input{sections/2relatedwork}

\input{sections/3method_new}
\input{sections/4experiments}

{\small
\bibliographystyle{ieee_fullname}
\bibliography{egbib}
}
\input{sections/appendix.tex}
\end{document}

%% file: sections/1introduction.tex
\section{Introduction}
Advancements in large-scale, high-quality datasets \cite{schuhmann2021laion,schuhmann2022laion5b,sharma2018conceptual} and generative diffusion models \cite{sohl2015deep, ho2020ddpm, rombach2022ldm} have significantly propelled image generation and editing \cite{shuai2024survey,wang2024genartist,yang2024mastering,wang2024divide,esser2024sd3}, producing synthetic digital artworks nearly indistinguishable from real creations. Extending this success to the temporal dimension, researchers have adapted diffusion models for video generation by treating videos as sequences of images over time. While this field has shown promising results, existing video generation and editing models \cite{chen2023videocrafter1,guo2023animatediff,ho2022vdm,wang2023lavie,ma2024latte,cong2023flatten} are typically trained on proprietary datasets with limited scale and diversity. This limitation confines their capabilities to specific scenarios and tasks, making it challenging to meet the diverse needs of users.

This is analogous to human society: no one can handle everything alone, but each person has unique strengths, and through collaboration, complex problems can be solved.
Similarly, while each open-source model may not be powerful enough individually to meet all demands, their collective potential is immense. This is because: \textbf{(\romannumeral 1)} \textbf{Specialized Capabilities:} Different models excel in specific tasks due to their specialized training and architectures\textemdash some generate videos from textual descriptions~\cite{ho2022vdm, ho2022imagenvideo, ma2024latte}, and some specialize in video editing~\cite{sun2024diffusion,wu2023tuneavideo,jeong2023groundavideo, feng2024ccedit,wang2023vid2vidzero}. Even within the same task, models may perform differently because of variations in their training datasets\textemdash some are better at generating natural scenery videos, while others excel in producing videos of animal movements. \textbf{(\romannumeral 2)} \textbf{Complex Task Handling:} By combining these models, even complex tasks can be accomplished; for example, integrating an image editing model~\cite{tumanyan2023pnp,brack2024ledits++} with an image-to-video model~\cite{sun2024diffusion, geyer2023tokenflow, cong2023flatten,wang2024videocomposer} enables users to edit an image and then convert it into a video.
Therefore, by effectively coordinating these open-source models, we can utilize them to accomplish a wide range of video generation and editing tasks.

However, manually coordinating these models~\cite{xing2023dynamicrafter,blattmann2023svd,chen2023seine,cong2023flatten,kara2024rave,geyer2023tokenflow,li2024vidtome,yuan2024mora} presents numerous challenges. First, users need to clarify their intentions and determine the specific tasks required to achieve them. For complex needs, users must analyze the problem and decompose it into multiple subtasks, allowing multiple models to collaborate. Second, users must have an in-depth understanding of each model's functionalities and applicability to select the most suitable model for each task. This process places high demands on users' expertise, increasing the barrier to entry and potentially leading to difficulties in finding satisfactory solutions promptly. Therefore, how to automatically coordinate these models to meet user requirements is an urgent problem that needs to be addressed.

To tackle these challenges and fully leverage the capabilities of existing models to meet diverse video generation and editing needs, we propose a novel Semantic Planning Agent (SPAgent) based video generation and editing system. The SPAgent integrates state-of-the-art open-source image/video generation and editing models as its tool library, where it automatically select appropriate model tool to generate or edit videos according to the users' intention.
Specifically, to enable SPAgent to perform effectively, we decompose this complex problem into three steps and allow SPAgent to proceed step by step to ensure quality results. First, SPAgent analyzes the user input to thoroughly understand the user's requirements. Next, it dynamically plans suitable execution routes based on the analyzed user intentions and some concluded principles. Then, for each task in every designed route, the agent selects appropriate models based on their capabilities to generate candidate videos for the user.
Furthermore, we enhance SPAgent's video quality assessment capability. With this ability, SPAgent can evaluate the quality of videos generated by new video generation or editing models, automatically acquiring capabilities information of these models. Consequently, the agent can autonomously expand its tool library with these new models, selecting them for tasks where they are most effective, thereby further enhancing the system's flexibility and practicality. 
Additionally, we manually annotate a multi-task generative video dataset, which will be open-sourced, to fine-tune the SPAgent for enhancing its aforementioned abilities. 
Experimental results demonstrate the effectiveness of the SPAgent in selecting appropriate models to meet diverse user needs, and its capability to autonomously expand its tool library.

%% file: sections/2relatedwork.tex
\section{Related Work}

\noindent \textbf{Video Generation Models.} With various diffusion models \cite{sohl2015deep, ho2020ddpm, rombach2022ldm} demonstrating remarkable capabilities on generating visual contents, diffusion model-based video generation models have achieved considerable development. Existing methods can be primarily categorized into text-to-video (T2V) and image-to-video (I2V) models based on the required input modalities. \textbf{T2V models} \cite{ho2022vdm, ho2022imagenvideo, ma2024latte, wang2023lavie, sora2024, opensora} are capable of generating realistic videos solely based on input text. The pioneering work Video Diffusion Models (VDM)~\cite{ho2022vdm} integrates factorized space-time attention and convolutional layers within the U-Net~\cite{cciccek20163dunet} framework for video generation. 
The following works such as LaVie~\cite{wang2023lavie} and VideoCrafter~\cite{chen2023videocrafter1, chen2024videocrafter2} leverage pre-trained T2I models as foundations to facilitate video generation. 
Moreover, several researches~\cite{chen2023gentron, ma2024latte, sora2024} explore DiT architecture for effective T2V generation. 
Although flexible and creative, T2V methods struggle with controllability due to the intricate expressiveness and comprehension challenges of the text modality. 
In light of this, some recent \textbf{I2V models} \cite{zhang2023i2vgen, blattmann2023svd, chen2023seine, xing2023dynamicrafter} use images as visual cues to enhance controllability. The image serves as a reference frame, with subsequent frames generated based on it, often collaboratively guided by text prompts. 
Although I2V models are capable of achieving more controllable video generation, the input images are not always available in practice.

\noindent \textbf{Video Editing Models.} Video-to-video (V2V) editing aims to modify the content of input video according to the given text prompt~\cite{sun2024diffusion,wu2023tuneavideo,jeong2023groundavideo, feng2024ccedit,wang2023vid2vidzero, geyer2023tokenflow, cong2023flatten,wang2024videocomposer}. Video editing is more challenging than image counterpart due to the need for temporal coherence between frames. Tune-A-Video~\cite{wu2023tuneavideo} extends the T2I model by adding a sparse causal attention module to achieve spatial-temporal relation modeling. To improve the temporal stability of the edited video, some works \cite{esser2023gen1, jeong2023groundavideo, feng2024ccedit, wang2024videocomposer, hu2023videocontrolnet} also attempt to explicitly introduce structured constraints like depth maps, motion and appearance conditions, \textit{etc}. Additionally, several V2V models employ attention feature injection \cite{liu2024videop2p, qi2023fatezero, wang2023vid2vidzero, geyer2023tokenflow, cong2023flatten} to edit videos through controlling hidden features. 

Although existing highly customized generation and editing models are performing increasingly well on different tasks, real-world user needs tend to be variable, making it challenging for a single model to fulfill all requirements. Thus exploring how to select the optimal model according to specific situations remains an important issue. 

\noindent \textbf{MLLM Agents.} In recent years, LLMs have shown remarkable emergent abilities~\cite{wei2022emergent} in text generation and reasoning. The advent of powerful LLMs like ChatGPT~\cite{openai2024chatgpt} and LLaMA~\cite{touvron2023llama, touvron2023llama2} has also boosted the development of multi-modal LLMs (MLLMs), \textit{e.g.,} GPT4~\cite{achiam2023gpt4}, LLaVA~\cite{liu2024llava}, InternVL~\cite{chen2024internvl}, which empower LLMs with the capability of visual reasoning. 
In the latest studies, MLLMs have been used as core components of AI agents to enable intelligent planning and decision-making, thereby addressing more complex multi-modal reasoning tasks, including game development~\cite{wu2023smartplay, koh2024visualwebarena}, mobile application operation~\cite{yang2023appagent, wang2024mobile}, embodied AI~\cite{wang2023describe, yang2024embodied}, \textit{etc}. Moreover, MLLM agents have also been used for image generation and editing \cite{chen2023llava-interactive, li2024mulan, wang2024divide, yang2024mastering, wang2024genartist}. 

Although MLLM agents are widely applied in image generation, research on their use in video production remains limited. 
Mora~\cite{yuan2024mora} unifies several video generation tasks in an agentic framework, however, it requires the user to select the desired task before generation and available tools for each task are limited. 
Our method integrates more diverse sets of generation and editing tools and is able to automate the entire video production process without human intervention.

%% file: sections/3method_new.tex
\begin{figure*}[tp]
    \centering
    \includegraphics[width=\linewidth]{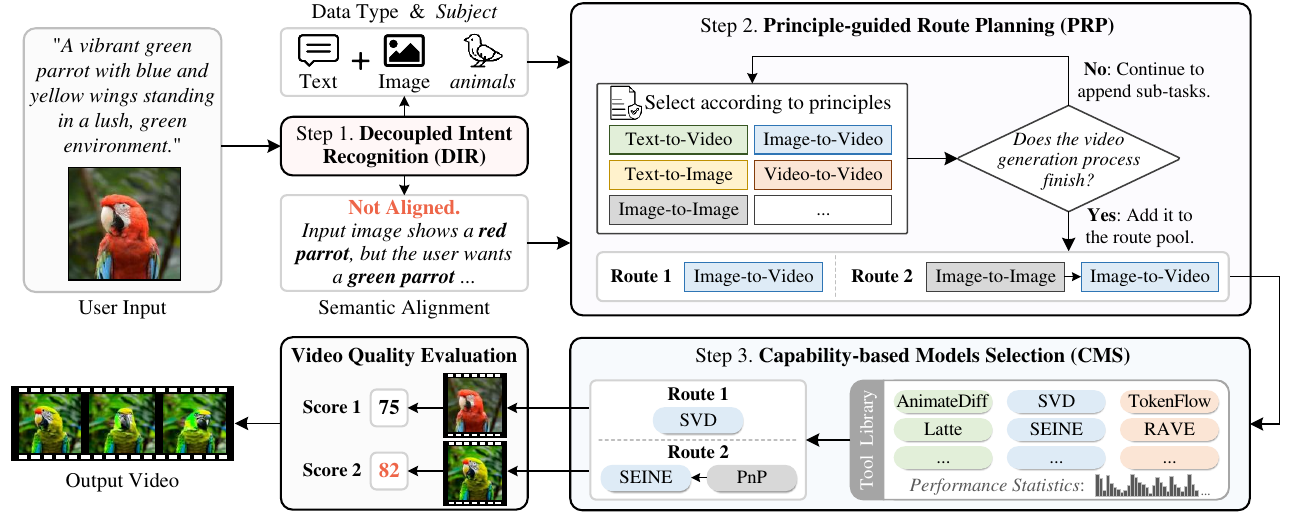}
    \caption{
        An illustration of our SPAgent. The MLLM agent acts as a coordinator. It elevates the multi-scenario video generative results by identifying the user intention, planning execution routes and models, and selecting the final output from all candidates.
    }
    \label{fig:workflow}
\end{figure*}

\section{Method}
\subsection{Motivation of Semantic Planning Agent}
We focus on unifying diverse video generation and editing capabilities in one system.
To achieve this goal, we consider various image and video generative models as tools, with the MLLM agent functioning as a central node that coordinates, selects, and evaluates these distributed tools.
Although substantial progress has been made in agent-coordinated text and image tasks, significant challenges remain unresolved in the video generation and editing domain:
\textbf{(\romannumeral 1)} \textbf{Accurately identifying the user intent} based directly on the input data is difficult. This is because the input data type varies and even the same type of input may require distinct workflows to meet user needs, which will be elaborated in \cref{subsubsec:3-2-1-intent}. 
\textbf{(\romannumeral 2)} \textbf{Complex user requirements} often necessitate sequential or combined execution of multiple tasks, with various models available for each task. It is quite challenging for the agent to select the best model combination in different situations without careful planning and profound awareness of the capabilities of different models. 
Therefore, to tackle the aforementioned challenges, we design a novel Semantic Planning Agent (SPAgent) system, thereby effectively achieving autonomous video generation and editing.

\subsection{Semantic Planning Agent}
\label{subsec:3-2-workflow}
As illustrated in the overall pipeline of our SPAgent in~\cref{fig:workflow}, the video generation process is decomposed into three key steps: (1) Decoupled Intent Recognition, (2) Principle-guided Route Planning, (3) Capability-based Models Selection. 
Additionally, the SPAgent is equipped with video quality evaluation capability. 
\subsubsection{Decoupled Intent Recognition (DIR)}
\label{subsubsec:3-2-1-intent}
Accurately identifying user intent poses significant challenges due to the diversity of input data types and the varied demands resulting from semantic misalignment between textual and visual inputs. Users may provide text, images, videos, or combinations thereof, each implying different intentions. For example, a text-only input typically requests text-to-video generation, while text accompanied by a video suggests video editing to align the content with the textual description. Even when input types are identical, differing semantic relationships can indicate distinct needs; semantically aligned text and image may imply animating the image into a video, whereas semantic discrepancies—such as the text describing ``green parrot'' while the image shows a ``red parrot''—might necessitate image editing (changing the parrot's color) before proceeding to video generation.

To address these challenges, we propose the Decoupled Intent Recognition (DIR) module that decomposes the complex intent recognition process into simpler sub-tasks. 
It is responsible for recognizing input data types and assessing the semantic alignment between textual and visual inputs.
By accurately interpreting both the input types and their semantic coherence, the agent can reliably identify user intent, enabling it to devise appropriate execution paths for subsequent stages. Additionally, this module will also predict the subject of the video that the user intends to use, which is adopted as guiding information for the agent in selecting the appropriate model tools, thereby enhancing the alignment of the generated videos with the user requirements. We formulate it as:
\begin{equation}
\begin{aligned}
\boldsymbol{E_T}, \boldsymbol{E_V}, \boldsymbol{A}, \boldsymbol{S}=\text{SPAgent}(\boldsymbol{D}_{uid}),
\end{aligned}
\label{eq:intention}
\end{equation}
where $\boldsymbol{D}_{uid}$ represents the user input data, which may include text, images, videos, or combinations thereof. $\boldsymbol{E_T} \in \{\text{`None'}, \text{`natural language text'}\}$ indicate the presence of text, where `None' indicates the absence of text input, and `natural language text' signifies its presence.
$\boldsymbol{E_V} \in \{\text{`None'}, \text{`single image'}\}, \text{`video'}\}$, where `None' no visual input, and `single image'  and `video' denote the type of visual input are image and video,respectively.
$\boldsymbol{A} \in \{\text{`Yes'}, \text{`No'}\}$ indicates whether the input text data and visual data are semantically aligned. $\boldsymbol{A} = \text{`Yes'}$ only if both text data and one type of visual data are present and semantically aligned; otherwise, $\boldsymbol{A} = \text{`No'}$. $\boldsymbol{S}$ represents the subject of the video the user intends to generate.
A detailed prompt for this module is provided in the Appendix.

\subsubsection{Principle-guided Route Planning (PRP)}
\label{subsubsec:3-2-2-route}
As the adage goes, ``\textit{If you fail to plan, you are planning to fail.}'' To effectively meet user requirements, we introduce the PRP module. This module assists the agent in planning an appropriate implementation route—an ordered sequence of sub-tasks—before selecting model tools for generation or editing tasks. Such route planning allows the agent to flexibly adapt to diverse user needs, greatly increasing the likelihood that the generated video aligns with user expectations.

\noindent \textbf{Prior Knowledge Injection.} To enable the agent to automatically plan a reasonable route based on user intent, we establish core route planning principles through analysis. For instance, when generating a video from text (\textit{i.e.}, text-only input), the possible routes include: 1) directly performing the T2V task, or 2) first executing the T2I task, followed by the I2V task to animate the image into a video. For video editing tasks (\textit{i.e.}, input includes text and video), the implementation route involves using the V2V task. When the input includes text and image, if they are well-aligned (indicating the user intends to animate the image into a video), the route should directly execute the I2V task; otherwise, the optimal route involves I2I task before I2V task. These principles are injected as prior knowledge into the agent to guide route planning.

\noindent \textbf{Iterative Route Planning.} With established principles, the agent generates the implementation route through the following steps: (1) Initial Task Selection: choosing the initial tasks based on input data type and semantic alignment; (2) Output Verification: confirming whether the task output is a video—if so, the route is complete; otherwise, proceeding to the next step; (3) Iterative Task Addition: determining the next task based on the current output type and user input, then returning to step (2) with the updated route. Furthermore, recognizing that some requests may have multiple viable routes while others have only one, the agent generates two candidate routes to ensure alternative methods are considered when possible. When only one route is applicable, two identical routes are provided for consistency. 
We formulate it as follows:
\begin{equation}
\begin{aligned}
\boldsymbol{R}_1, \boldsymbol{R}_2=\text{SPAgent}(\boldsymbol{D}_{uid}, \boldsymbol{E_T}, \boldsymbol{E_V}, \boldsymbol{A}),
\end{aligned}
\end{equation}
where $\boldsymbol{R}_1, \boldsymbol{R}_2$ denote the two predicted route lists.

\begin{table}[t]
\centering
\caption{Main models in the tool library of our SPAgent.}
\label{tab:tool-library}
\resizebox{\columnwidth}{!}{%
\begin{tabular}{@{}lccc@{}}
\toprule
 Task & Text-to-Video & Video-to-Video & Image-to-Video \\ \midrule
\multirow{4}{*}{Tools} & LATTE~\cite{ma2024latte} & TokenFlow~\cite{geyer2023tokenflow} & SVD~\cite{blattmann2023svd} \\
 & LaVie~\cite{wang2023lavie} & VidToMe~\cite{li2024vidtome} & Seine~\cite{chen2023seine} \\
 & AnimateDiff~\cite{guo2023animatediff} & RAVE~\cite{kara2024rave} & DynamiCrafter~\cite{xing2023dynamicrafter} \\
 & OpenSora~\cite{opensora} & FLATTEN~\cite{cong2023flatten} &  \\ \bottomrule
\end{tabular}%
}
\end{table}

\subsubsection{Capability-based Models Selection (CMS)}

\noindent \textbf{Tool Library.} Upon confirming the implementation routes, the agent will select appropriate models from the tool library to perform each task in these routes. Our tool library supports tasks such as text-to-video generation, video-to-video editing, and image-to-video generation by integrating recent open-source models (see~\cref{tab:tool-library}). Additionally, we adopt SD3~\cite{esser2024sd3} and PnP~\cite{tumanyan2023pnp} as default tools for text-to-image generation and image-to-image editing, respectively.

\noindent \textbf{Model Selection.} Recognizing that distinct models excel in different subjects for video generation or editing, we evaluated and categorized each model’s strengths within the tool library (refer to~\cref{subsec:3-4-dataset} for details). These statistical data serve as fine-grained capabilities information of candidate models, enabling the agent to more precisely match user input with the most relevant model tools. During model selection, the agent utilizes user input, subject information extracted by the DIR module, and model performance statistics to select the optimal model for each task in the designed routes, invoking the chosen model(s) to generate the videos. The prompt design for this module is provided in the Appendix, and is simply formulated as follows:
\begin{equation}
\begin{aligned}
\boldsymbol{M}_1, \boldsymbol{M}_2=\text{SPAgent}(\boldsymbol{D}_{uid}, \boldsymbol{D}_{m}, \boldsymbol{R}_1, \boldsymbol{R}_2, \boldsymbol{E_T}, \boldsymbol{E_V}, \boldsymbol{S}),
\end{aligned}
\end{equation}
where $\boldsymbol{D}_{m}$ denotes the model performance statistics, i.e., the information about the capabilities of each model in the tool library across various subjects.
$\boldsymbol{M}_1, \boldsymbol{M}_2$ denote the predicted model lists for the route lists $\boldsymbol{R}_1, \boldsymbol{R}_2$, respectively.

\subsubsection{Video Quality Evaluation}
In the CMS step, our agent requires $\boldsymbol{D}_{m}$ as input, which encapsulates the capability information of each model in the tool library across various subjects. In the training phase, such information is derived from human-annotated scores of the videos generated by each model in the tool library. During the inference phase, when we aim to introduce a new model into the library for the agent to select from, it is necessary to acquire this model's capability information and update it into $\boldsymbol{D}_{m}$, which is labor-intensive and time-consuming for human annotating.

To address this, we fine-tune the SPAgent's video quality evaluation capabilities using a carefully designed video quality evaluation prompt. With this enhancement, the agent can autonomously assess the quality of videos produced by new video generation or editing models, automatically acquiring performance information about these models. Consequently, the agent can automatically expand its tool library by incorporating these models without manual intervention. Specifically, in the designed prompt, the agent evaluates videos based on intrinsic quality and alignment with input data: intrinsic quality assesses visual aesthetics and physical plausibility, ensuring the video appears natural and realistic, while alignment quality measures the match between video content and the user's specified input in both visual and motion aspects. By integrating these evaluations, the agent produces a final score that reflects overall quality, ensuring high scores indicate outputs that are both visually compelling and aligned with user intent.

\begin{figure}[tp]
    \centering
    \includegraphics[width=\linewidth]{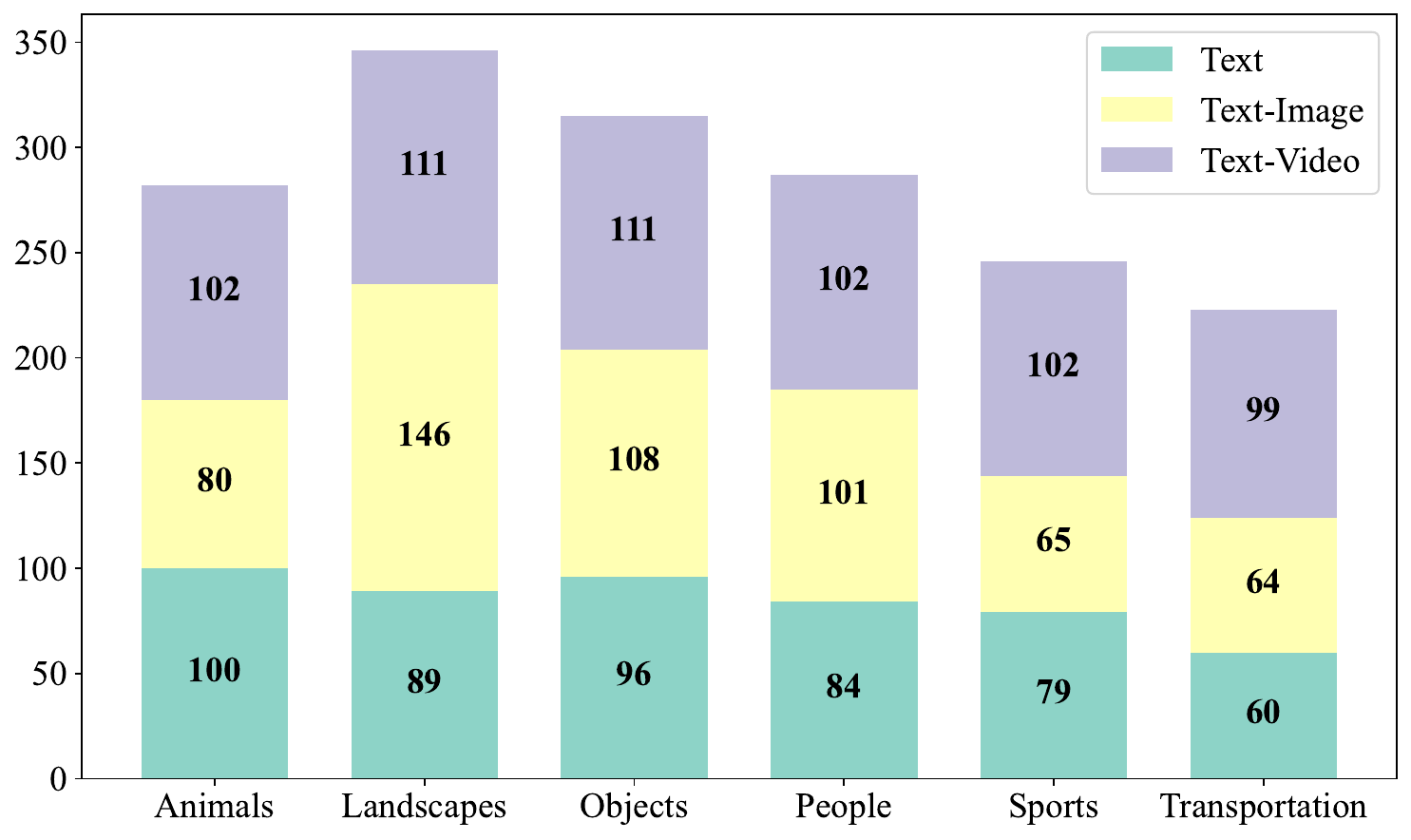}
    \caption{Data statistics of our dataset in different categories.}
    \label{fig:input-data}
\end{figure}

\subsection{Dataset Curation}
\label{subsec:3-4-dataset}
With above workflow, our agent is able to automate the video generation and editing, but it often falls short of producing optimal results. 
To enhance the capabilities of SPAgent, Supervised Fine-Tuning (SFT) is a feasible solution. 
Considering existing datasets are typically designed for single tasks, we construct a comprehensive multi-task generative video dataset, using multi-modal and multi-purpose inputs. 

\noindent \textbf{Input Data Collection.} The input data of our framework involves texts, images and videos. Since existing generative models often show varying performance across themes, we collect the input data from six distinct categories: \texttt{sports}, \texttt{transportation}, \texttt{people}, \texttt{animals}, \texttt{objects}, and \texttt{landscapes}. 
Textual samples originate from three primary avenues: AI-generated text (via ChatGPT), online content, and textual descriptions from existing datasets.
We eventually select a total of 508 texts and the length of each sentence is between 2 and 34. 
For image and video data, we collect open-licensed content from the internet and existing datasets. Each image or video is annotated with one descriptive prompt for generation tasks and three editing prompts for editing tasks. Consistent with common practice, these editing prompts modify the foreground subject, background, or style of the visual input. 
The statistics of our collected input data are shown in Fig.~\ref{fig:input-data}.

\noindent \textbf{Output Data Annotation.} 
We manually define the task and select feasible tools, which may not be optimal, to generate the output video. We then assign an opinion score to each output video, considering two primary perspectives: 1) Intrinsic Quality primarily evaluates whether the generated video appears natural and coherent, and 2) Prompt Alignment focuses on verifying whether the generated content and temporal changes of the output video align with the input prompt text. 
Each output is assessed by three annotators, and the subjective opinion scores are averaged to mitigate personal bias, resulting in a final Mean Opinion Score (MOS). 
Finally, we accumulated a total of 7712 generated video MOS annotations, offering greater diversity in content, tasks, and dimensions compared to existing benchmarks.

%% file: sections/4experiments.tex
\section{Experiments}
\subsection{Setup}

\noindent \textbf{Dataset Split}. We manually split the curated dataset into training set and testing set for SFT and evaluation, respectively. To be precise, we divide the dataset by employing stratified sampling across the six categories, assigning 90\% to the training set and 10\% to the testing set. 
\noindent \textbf{Baselines and Evaluation}. To evaluate the performance of our system, we selected the following comparison methods: (1) Tool Models within the system; (2) Mora~\cite{yuan2024mora}; (3) GPT-4o~\cite{hurst2024gpt-4o} which is directly used as agent to accomplish each task; (4) Ours$^{\ast}$ which integrates the GPT-4o into our three step agent framework to accomplish each task. Besides, our SPAgent employs the open-source InternLM-xComposer-2.5~\cite{zhang2024internlm} as the MLLM, which is further fine-tuned on our constructed training prompts using SFT.

\begin{table*}
\centering
\caption{The MOS results and task completion rates (in \textcolor{cvprblue}{blue} font) of our SPAgent and baselines. }
\resizebox{\textwidth}{!}{
\begin{tabular}{c|cccc|c|c|c|c}
    \toprule
    \textbf{Task} & \multicolumn{4}{c|}{\textbf{Tool Models within the system}} 
        & \textbf{Mora}~\cite{yuan2024mora}
        & \textbf{GPT-4o}~\cite{hurst2024gpt-4o} & \textbf{Ours$^{\ast}$} & \textbf{Ours} \\
    \midrule
    \multirow{2}{*}{{Text-to-Video}} 
        & \textit{AnimateDiff}~\cite{guo2023animatediff}
        & \textit{LaVie}~\cite{wang2023lavie}
        & \textit{Latte}~\cite{ma2024latte}
        & \textit{OpenSora}~\cite{opensora}
        &\multicolumn{1}{c|}{---}
        & (\textcolor{cvprblue}{$95.3\%$})
        & (\textcolor{cvprblue}{$93.8\%$})
        & (\textcolor{cvprblue}{$100.0\%$})
        \\
        & 10.497 
        & 9.871
        & 9.801 
        & 9.006 
        & 9.737
        & 11.636 
        & 12.133 
        & 12.386 
        \\
    \midrule
    \multirow{2}{*}{{Image-to-Video}} 
        & \textit{DynamiCrafter}~\cite{xing2023dynamicrafter}
        & \textit{SEINE}~\cite{chen2023seine}
        & \textit{SVD}~\cite{blattmann2023svd}
        &\multicolumn{1}{c|}{---}
        & \multicolumn{1}{c|}{---}
        & (\textcolor{cvprblue}{$87.7\%$})
        & (\textcolor{cvprblue}{$98.2\%$})
        & (\textcolor{cvprblue}{$100.0\%$})
        \\ 
        & 12.974 
        & 12.453 
        & 13.740
        & \multicolumn{1}{c|}{---}
        & 11.870
        & 12.924 
        & 14.181 
        & 14.406 
        \\
    \midrule
    \multirow{2}{*}{{Video-to-Video}} 
        & \textit{FLATTEN}~\cite{cong2023flatten} & \textit{RAVE}~\cite{kara2024rave} & \textit{TokenFlow}~\cite{geyer2023tokenflow} & \textit{VidToMe}~\cite{li2024vidtome} 
        &\multicolumn{1}{c|}{---}
        & (\textcolor{cvprblue}{$40.8\%$})
        & (\textcolor{cvprblue}{$95.9\%$})
        & (\textcolor{cvprblue}{$91.8\%$})
        \\
        & 9.864 
        & 11.088 
        & 9.673 
        & 9.966 
        & 7.767
        & 7.734 
        & 11.121 
        & 10.692 
        \\
    \bottomrule
\end{tabular}}
\label{tab:mos}
\end{table*}

\begin{table*}
\centering
\caption{Comparison of our SPAgent and baseline methods in generated video quality evaluation capabilities.}
\resizebox{\textwidth}{!}{
\begin{tabular}{c|c|ccc|ccc|ccc}
    \toprule
    \multirow{2}{*}{\textbf{Category}}
    & \multirow{2}{*}{\textbf{Method}}
        & \multicolumn{3}{c|}{\textbf{Text-to-Video}} 
        & \multicolumn{3}{c|}{\textbf{Image-to-Video}}
        & \multicolumn{3}{c}{\textbf{Video-to-Video}} \\
    
      & & SRCC $\uparrow$ & PLCC $\uparrow$ & KRCC $\uparrow$
        & SRCC $\uparrow$ & PLCC $\uparrow$ & KRCC $\uparrow$
        & SRCC $\uparrow$ & PLCC $\uparrow$ & KRCC $\uparrow$ \\
    \midrule
    
    \multirow{4}{*}{\makecell{\textbf{Feature-based}\\\textbf{Methods}}}
    & CLIP Frame~\cite{radford2021learning} 
        & 43.71 & 46.58 & 30.63
        & \best{61.65} & \best{66.71} & \secondbest{45.48}
        & 6.73 & 17.22 & 4.69 \\
    & DINO Frame \cite{caron2021emerging}
        & 37.42 & 43.09 & 26.02
        & 46.83 & 55.83 & 33.30
        & 12.07 & 27.55 & 8.38 \\
    & CLIP Text \cite{radford2021learning} 
        & 19.62 & 20.86 & 13.19
        & 21.21 & 22.03 & 14.26
        & 14.09 & \secondbest{26.73} & 10.36\\
    & ViCLIP Text \cite{wanginternvid}
        & 38.09 & 40.89 & 26.44
        & 5.93 & 8.80 & 4.20
        & \secondbest{19.03} & 25.37 & \secondbest{13.06} \\
    \midrule
    \multirow{6}{*}{\makecell{\textbf{VQA-based}\\\textbf{Methods}}}
    & Aesthetic \cite{schuhmann2021laion}
        & 40.02 & 42.72 & 28.18
        & 48.07 & 45.77 & 32.93
        & 6.42 & 13.83 & 4.07\\
    & DOVER \cite{wu2023exploring}
        & 17.57 & 19.43 & 12.16
        & 24.87 & 26.24 & 17.05
        & 1.99 & 13.09 & 1.53\\
    & VS Quality \cite{he2024videoscore}
        & 49.27 & 49.91 & 34.80
        & 19.96 & 48.30 & 14.49
        & 12.26 & 18.19 & 8.72 \\
    & VS Consistency \cite{he2024videoscore}
        & \secondbest{52.78} & \secondbest{54.73} & \secondbest{37.22}
        & 29.41 & 53.62 & 21.46
        & 9.97 & 17.77 & 6.45 \\
    & VS Factuality \cite{he2024videoscore}
        & 51.41 & 52.80 & 36.39
        & 25.42 & 51.55 & 18.69
        & 13.56 & 20.75 & 9.18 \\
    \cmidrule{2-11}
    & Ours
        & \best{57.17} & \best{60.41} & \best{45.81}
        & \secondbest{60.40} & \secondbest{64.18} & \best{48.02}
        & \best{44.24} & \best{47.04} & \best{33.49}\\
    \bottomrule
\end{tabular}}
\label{tab:vqa-all}
\end{table*}

\subsection{Experimental Results}
To evaluate the effectiveness of our SPAgent in coordinating models for video generation and editing tasks, we compared the quality of videos produced by our SPAgent with those generated by baseline methods, using Mean Opinion Score (MOS) as the quality metric. The experimental results are shown in \cref{tab:mos}. Notably, GPT-4o, Ours$^{\ast}$, and Ours rely on automatic model selection, so the task completion rate is reported in blue font as an additional metric. Higher values indicate a stronger capability to choose the correct model for successfully accomplishing the corresponding task. 

From the results in \cref{tab:mos}, we can derive the following insights: (1) Our SPAgent achieves higher overall video generation and editing quality. For instance, in the T2V task, the MOS of videos generated by models selected by our SPAgent reaches 12.386, surpassing any individual tool in the library. This demonstrates the agent's capability to accurately analyze user input and select the most suitable model, resulting in higher-quality video outputs.
(2) The agent exhibits a strong advantage in accurately selecting models that can correctly fulfill user-intended tasks. Specifically, it achieves a 100\% task completion rate for both T2V and I2V tasks, and maintains a high accuracy rate in V2V tasks, comparable to GPT-4o. This indicates that the agent effectively interprets user intent and reliably selects the correct model for each task.
(3) Compared to GPT-4o, Ours$^{\ast}$ demonstrates notable improvements in both model selection accuracy and generated video quality. This suggests that our workflow enhances the ability of MLLMs to decompose the complex problem of automatic model selection for video generation, enabling better user intent understanding and model selection to fulfill the required tasks.

\subsection{Video Quality Assessment Capability}

To assess the agent's evaluation performance, we conducted validations on three tasks: text-to-video, image-to-video, and Video-to-Video.
We employed widely-used statistical metrics—Spearman correlation (SRCC), Pearson correlation (PLCC), and Kendall correlation (KRCC)—to measure the consistency between the agent's evaluations and human judgments. Higher values of these correlation coefficients indicate a stronger alignment with human assessments, reflecting a better evaluation ability. We compare our SPAgent with four feature-based methods: CLIP Frame~\cite{radford2021learning}, DINO Frame \cite{caron2021emerging}, CLIP Text \cite{radford2021learning}, ViCLIP Text \cite{wanginternvid}; and three QA model-based methods: Aesthetic \cite{schuhmann2021laion}, DOVER \cite{wu2023exploring},  VideoScore \cite{he2024videoscore}. As VideoScore evaluates five different dimensions, we present only the best three dimensions\textemdash Quality, Consistency, and Factuality\textemdash in our results due to space limitations, denoted individually as 'VS $dimension$'.

As shown in~\cref{tab:vqa-all}, our agent achieved superior results in the text-to-video and video editing tasks. For instance, in the text-to-video task, the SRCC between our SPAgent's evaluations and human scores reached 57.17, outperforming the best competing model, VS Consistency, which achieved 52.78—a relative improvement of 8.3\%. 
In the image-to-video task, SPAgent performed comparably to the leading method, CLIP Frame. While it lagged slightly behind in SRCC and PLCC, it achieved the best evaluation effect in terms of the KRCC.
In summary, SPAgent attains the best or near-best evaluation performance across all comparing methods, which demonstrates our agent is also capable of effectively assessing the quality of generated videos.

\begin{table*}
\centering
\caption{Results of ablation study on the components of our proposed SPAgent.}
\vspace{-0.1cm}
\resizebox{\textwidth}{!}{
\begin{tabular}{c|c|c|c|c|c}
    \toprule
        & \multicolumn{1}{c|}{\textbf{Text-to-Video}} 
        & \multicolumn{1}{c|}{\textbf{Image-Edit-to-Video}} 
        & \multicolumn{1}{c|}{\textbf{Image-to-Video}}
        & \multicolumn{1}{c|}{\textbf{Video-to-Video}}
        & \multicolumn{1}{c}{\textbf{Average Results}} \\
    \midrule
    One-Step 
        & 39.47 
        & \secondbest{65.63} 
        & 12.50 
        & 38.78 
        & 48.48 \\
    w/o Data Analysis 
        & 47.39
        & 57.22
        & 37.49
        & 49.48
        & 51.11 \\
    w/o Route Planning 
        & 43.86
        & 63.02 
        & 59.38
        & 44.90
        & 56.91 \\
    w/o Principle 
        & \secondbest{47.81}
        & 56.25
        & 63.41
        & 45.41
        & \secondbest{57.25} \\
    w/o Subject 
        & 45.61 
        & 58.07  
        & \secondbest{64.06}
        & \best{52.04}
        & 56.35  \\
    \midrule
    Our model 
        & \best{48.25} 
        & \best{66.15} 
        & \best{64.84}
        & \secondbest{48.99}
        & \best{60.77}  \\
    \bottomrule
\end{tabular}}
\vspace{-0.3cm}
\label{tab:ablation}
\end{table*}

\begin{figure}[tp]
    \centering
    \includegraphics[width=\linewidth]{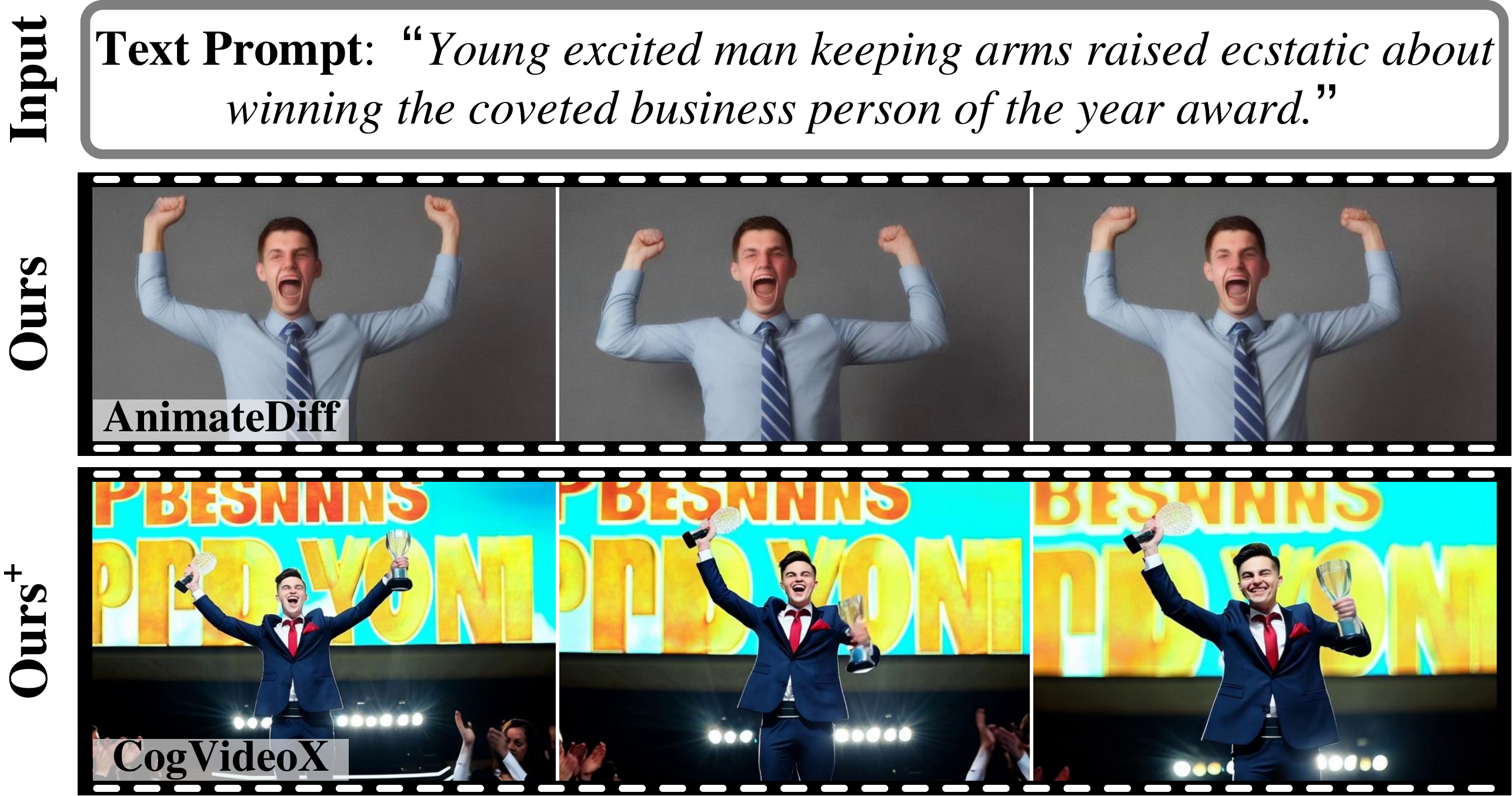}
    \vspace{-0.3cm}
    \caption{
        Comparison of output videos generated by SPAgent with and without integrating CogVideoX into its tool library
    }
    \vspace{-0.5cm}
    \label{fig:zero-shot}
\end{figure}

\subsection{Tool Library Automatic Expansion}
To validate SPAgent's capability to automatically expand its tool library, we conduct a dedicated experiment on the text-to-video generation task. During the training phase, the T2V model CogVideoX~\cite{yang2024cogvideox} was intentionally not included in the tool library. Thus, we conduct the experiment to verify whether the agent can autonomously incorporate CogVideoX into its tool library and select it in subsequent appropriate scenarios. Specifically, we first use the agent to evaluate the quality of videos generated by CogVideoX, obtaining performance metrics across different subjects through statistical analysis. Then these information will be fed into the agent together with the information of other models.

As shown in in \cref{fig:zero-shot}, the row labeled ``Ours'' displays the video outputs generated by our agent before integrating CogVideoX, while ``Ours$^{+}$'' shows the results after the agent automatically added CogVideoX without any human-annotated video quality information. It can be observed that when the user input aligns with CogVideoX's capabilities, the agent autonomously selects this model, significantly improving video quality. These findings verified the scalability of our agent, which can effectively evaluate and integrate new models without human intervention.

\begin{figure*}[tp]
    \centering
    \includegraphics[width=\textwidth]{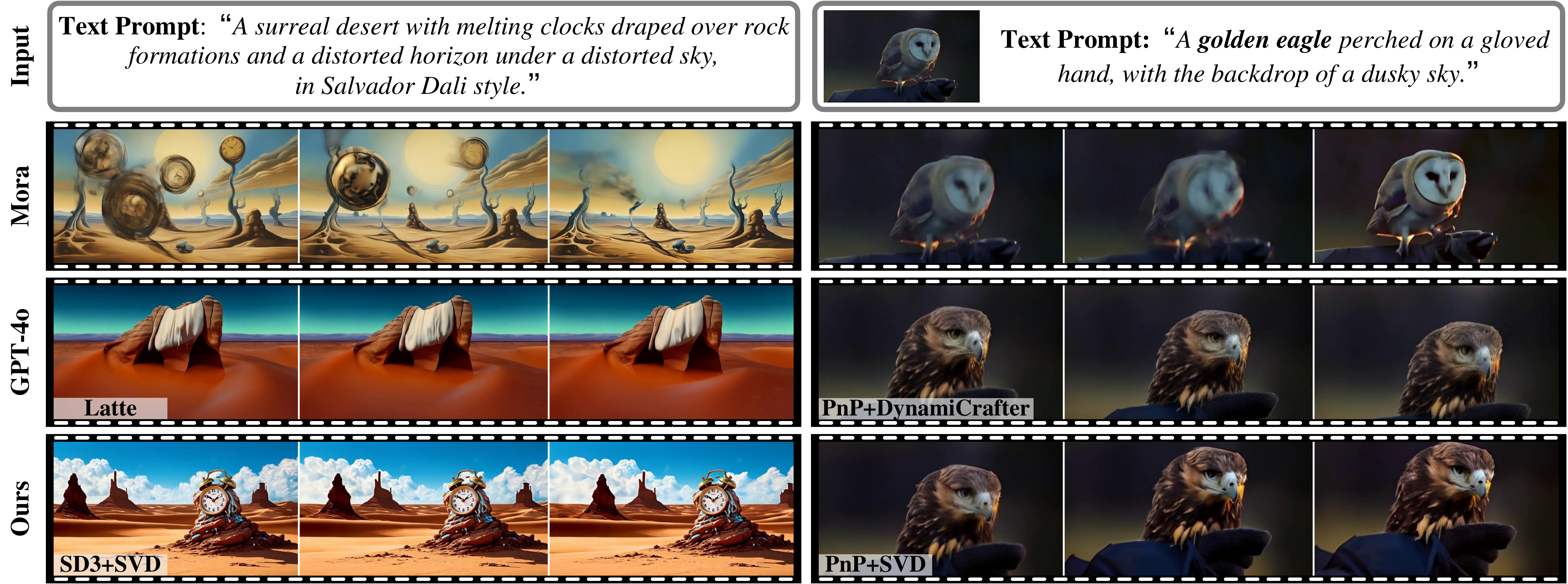}
    \caption{
        Comparison of the output videos generated by different methods given various user input data types and requirements.
    }
    \label{fig:t2v+i2v}
\end{figure*}

\subsection{Ablation Study}
To evaluate the impact of each component in our proposed SPAgent, we conducted ablation studies by designing five variants. (1) \textbf{One-Step}: This variant omits the DIR module and PRP module, directly predicting the list of models without user intent recognition and execution route planning steps. (2) \textbf{w/o Data Analysis}: Without DIR module, this variant skips the process of analyzing input data to identify user intent, while remains the rest steps. (3) \textbf{w/o Route Planning}: This one removes the PRP module and predicts execution models right after the DIR module. (4) \textbf{w/o Principle}: In this variant, we do not incorporate the planning principles during the route planning step, while the other steps remain consistent with the full method. (5) \textbf{w/o Subject}: This variant disregards the subject information of user's desired content to perform the CMS step. 

In this experiment, we use the suitable model list selection precision as the evaluation metric. When the video generated by the model list selected by the agent ranks among the top two in quality compared to videos generated by all candidate model lists, it is considered a positive selection; otherwise, it is a negative one. The precision is then defined as the ratio of positive model lists, with higher values indicating better selection performance by the agent.

\cref{tab:ablation} presents the evaluation results of our SPAgent and its ablation variants across four tasks: Text-to-Video (T2V), Image-Edit-to-Video (IE2V), Image-to-Video (I2V), and Video-to-Video (V2V). Our SPAgent consistently outperforms all its variants, underscoring the effectiveness of each component in SPAgent.
\textbf{(1)} The \textbf{One-Step} variant performs the worst, achieving an accuracy of only 12.50\% in the I2V task and 39.47\% in the T2V task. This indicates that lacking task decomposition and directly predicting the required models leads to suboptimal results, confirming the importance of multi-step task execution for high-quality video generation.
\textbf{(2)} The \textbf{w/o Data Analysis} variant exhibits significant performance drops across all tasks, especially in the IE2V and I2V that require deep intent understanding. This emphasizes the critical role of user intent analysis in handling tasks with multi-modal inputs.
\textbf{(3)} Our model outperforms the \textbf{w/o Route Planning} variant in all tasks (average accuracy of 60.77\% vs. 56.91\%). 
This demonstrates that planning the execution route first and then selecting appropriate models based on the route can effectively improve the accuracy of model selection.
\textbf{(4)} The \textbf{w/o Principle} variant shows a significant performance drop in most tasks, especially in the IE2V task where it achieves only 56.25\% accuracy compared to the full model's 66.15\%. This highlights the value of incorporating predefined principles during the route planning stage to ensure optimal task execution strategies.
\textbf{(5)} The \textbf{w/o Subject} variant reveals the importance of considering subject information during model selection. The lack of subject awareness leads to lower scores in the T2V (45.61\%) and V2V (52.04\%) tasks, indicating that subject-specific model matching positively impacts the quality of generated results.

\subsection{Qualitative Analysis}
To intuitively analyze the effectiveness of our method, we compare the videos produced by Mora~\cite{yuan2024mora}, GPT-4o and our SPAgent given different types of user input. 

\noindent \textbf{High-quality generation.} When only text input is given, \textit{i.e.,} text-to-video generation task, Mora struggles to produce ideal videos with relative complex text prompts. Although GPT-4o can determine the right routes to complete the T2V task in many cases, it is challenging to choose the optimal model. In contrast, the videos generated by our method show better visual aesthetics and prompt alignment. For instance, on the left side of~\cref{fig:t2v+i2v}, our produced video depicts a more accurate concept of the 'melting clock', which demonstrates our model is able to effectively identify the subject of the user input text, thus selecting the most suitable model. 

\noindent \textbf{Autonomous editing.} When the user provides both image (or video) and text input, our model will autonomously judge the semantic alignment between them, so that incorporating necessary editing procedures during video generation. As shown in the right side of~\cref{fig:t2v+i2v}, the user inputs an image of a barn owl with a text prompt aiming to change it into a golden eagle. 
Our model and GPT-4o successfully identified the editing requirements of the user so that edited the main entity of the output video into a golden eagle, while Mora failed to meet the editing needs of the user.

\section{Conclusion}
We propose the Semantic Planning Agent (SPAgent) for building video generation and editing systems. SPAgent analyzes user input and, through Decoupled Intent Recognition, Principle-guided Route Planning, and Execution Model Selection, automatically coordinates vision generation and editing models in the tool library to accomplish user tasks. Moreover, we fine-tuned SPAgent's video quality evaluation capability, enabling it to automatically assess new models' performance and expand the tool library. To achieve this, we manually annotated the quality of videos generated by models in the tool library to construct a training dataset, which will be publicly available. Experimental results demonstrate the effectiveness of the SPAgent in coordinating models to generate or edit videos, highlighting its potential to meet diverse user needs and autonomously expand its tool library.

%% file: sections/appendix.tex
\clearpage
\setcounter{page}{1}
\maketitlesupplementary

\appendix

\section{Prompt Templates for SPAgent}

\begin{figure}[ht]
\scriptsize
\centering
\begin{tcolorbox}
\textbf{\# Your job}\\
As an expert in interpreting user inputs, your task is to identify the type of the `user\_input\_text' and `user\_input\_image', and determine the subject of the input data.\\
\\
\textbf{\# Input Data}\\
The user\_input\_text: \{text\_prompt\}. \\
The user\_input\_image: \{image\_prompt\}. \\
\\
\textbf{\# Data type determining pipeline}\\
Step1: Identify the type of `user\_input\_text' as either ``None", or natural language text.\\
Step2: Identify the type of `user\_input\_image' as either ``None", a single image, or a video.\\
\text{\quad} - If the `user\_input\_image' lacks visual content, classify it as ``None".\\
\text{\quad} - If the `user\_input\_image' is an image composed of multiple pictures, classify it as video; \\
\text{\quad} - Otherwise, classify it as single image.\\
Step3: Identify the subject of the input data.\\
\text{\quad} - Based on the content of `user\_input\_text', select the subject from one of the six categories: sports, transportation, people, animals, objects, or landscapes.\\
\\
\textbf{\# Output Template}\\
Repalce Variable in `\{\{\}\}'\\
Please generate the types of the `user\_input\_text' and `user\_input\_image' and the subject of these input data:\\
1. Type of user\_input\_text: \{\{type1\}\}.\\
2. Type of user\_input\_image: \{\{type2\}\}.\\
3. The subject: \{\{name\}\}.
\end{tcolorbox}
\caption{Prompt template for input datatype and subject prediction in DIR.}
\label{fig:datatype_subject}
\end{figure}

\begin{figure}[ht]
\scriptsize
\centering
\begin{tcolorbox}
\textbf{\# Your job}\\
As an expert in evaluating semantic alignment, your role is to determine whether the user\_input\_text and user\_input\_image are aligned-well.\\
\\
\textbf{\# Input Data}\\
`user\_input\_text': \{text\_prompt\}.\\
`user\_input\_image': \{image\_prompt\}.\\
\\
\textbf{\# Alignment determing pipeline}\\
Step1: Assess whether the content of user\_input\_text and user\_input\_image are semantically aligned.\\
\\
\textbf{\# Output Template}\\
Repalce Variable in `\{\{\}\}'\\
Please provide the result indicating whether the two data are semantically aligned-well:\\
Aligned: \{\{answer\}\}.
\end{tcolorbox}
\caption{Prompt template for input alignment prediction in DIR.}
\label{fig:aligment}
\end{figure}

\begin{figure}[ht]
\scriptsize
\centering
\begin{tcolorbox}
\textbf{\# Your job}\\
As an expert in design routes to accomplish user intentions, your job is to determine two possible route\_lists for generating a video based on the `user\_input\_text' and `user\_input\_image'.\\
\\
\textbf{\# Definitions of route\_list.}\\
Each route\_list is a sequence of tasks designed to generate a video that satisfies the user's requirements. Each task in each route\_list must be one of the following five tasks: text\_to\_video, video\_to\_video, image\_to\_video, text\_to\_image, or image\_to\_image.\\
1. text\_to\_video: Generates a video based on the `user\_input\_text'. \\
2. video\_to\_video: Edits the `user\_input\_image' (a video) into the target video according to the `user\_input\_text'\\
3. image\_to\_video: Animates the `user\_input\_image' into a video. If `user\_input\_text' is provided, it must align well with the input image, and the generated video should be semantically consistent with both.\\
4. text\_to\_image: Generates an image based on the `user\_input\_text'.\\
5. image\_to\_image: Edits the `user\_input\_image' into the target image according to the `user\_input\_text'.\\
\\
\textbf{\# Input Data}\\
`user\_input\_text': \{text\_prompt\}.\\
Type of `user\_input\_text': \{type\_text\}.\\
`user\_input\_image': \{image\_prompt\}.\\
Type of `user\_input\_image': \{type\_image\}.\\
Value of `aligned': \{alignwell\}. \\
\\
Generally, there are five possible route\_lists: [text\_to\_video], [text\_to\_image, image\_to\_video], [video\_to\_video], [image\_to\_video], [image\_to\_image, image\_to\_video].\\
There are some principles to design the route\_lists:\\
1. When the type of `user\_input\_image' is `None', the two possible route\_lists can be either [text\_to\_video] or [text\_to\_image, image\_to\_video].\\
2. When the type of `user\_input\_image' is a video, both selected routes should be [video\_to\_video].\\
3. When the type of `user\_input\_image' is a single image:\\
\text{\quad}- If `aligned' is `No', the two selected routes should be [image\_to\_image, image\_to\_video].\\
\text{\quad}- If `aligned' is `Yes', the two selected routes should be [image\_to\_video].\\
\\
\textbf{\# Route\_lists selection pipeline}\\
Step1: Generate the first potential route list:\\
\text{\quad}Step1.1: Select the initial task for the route list based on the type of `user\_input\_image', the `aligned' value, and the established principles.\\
\text{\quad}Step1.2: Check if the content generated by the previously selected task(s) in the route is a video. If it is a video, the route is complete. If not, proceed by selecting the next task according to the type of the generated content and input data information. \\
\text{\quad}Step1.3: Repeat Step1.2 until the content generated by the selected tasks in the route is a video.\\
Step2: Use the same process as in Step 1 to generate the second potential route list.\\
\\
\textbf{\# Output Template}\\
Repalce Variable in `\{\{\}\}'\\
Please generate the two selected route\_lists based on following template:
Route\_lists: [\{\{task 1\}\}, ...], [\{\{task 1\}\}, ...] 
\end{tcolorbox}
\caption{Prompt template for PRP.}
\label{fig:route}
\end{figure}

\begin{figure}[ht]
\scriptsize
\centering
\begin{tcolorbox}
\textbf{\# Your job}\\
As an expert in identifying user intentions, your task is to predict the list of models that can execute the tasks specified in each route\_list.\\
\\
\textbf{\# Definitions of tasks}\\
There are five task:
1. text\_to\_video: Generates a video based on the `user\_input\_text'. \\
2. video\_to\_video: Edits the `user\_input\_image' (a video) into the target video according to the `user\_input\_text'\\
3. image\_to\_video: Animates the `user\_input\_image' into a video. If `user\_input\_text' is provided, it must align well with the input image, and the generated video should be semantically consistent with both.\\
4. text\_to\_image: Generates an image based on the `user\_input\_text'.\\
5. image\_to\_image: Edits the `user\_input\_image' into the target image according to the `user\_input\_text'.\\
\\
\textbf{\# Models for each task}\\
Model information is provided as `(model\_name, strength\_dict)', where `model\_name' is the name of the model, and `strength\_dict' is a dictionary with key-value pairs indicating the model's strengths. Each key represents a subject of the main object in the generated video, and the value (ranging from 100 to 2000, with 100 being the lowest and 2000 being the highest) reflects the model's capability in generating videos for that subject.\\
Candidate model(s) for text\_to\_video task: \{model\_prompt\_t2v\}.\\
Candidate model(s) for video\_to\_video task: \{model\_prompt\_v2v\}.\\
Candidate model(s) for image\_to\_video task: \{model\_prompt\_i2v\}.\\
Candidate model(s) for text\_to\_image task: \{model\_prompt\_t2i\}.\\
Candidate model(s) for image\_to\_image task: \{model\_prompt\_i2i\}.\\
\\
\textbf{\# Input Data}\\
`input\_subject': \{prompt\_subject\}.\\
`user\_input\_text': \{text\_prompt\}.\\
`user\_input\_image': \{image\_prompt\}. \\
`input\_route\_lists': \{routelist\_prompt\}.\\
\\
\textbf{\# Model(s) selection}\\
Step1: For the first route\_list in the `input\_route\_lists', select the best model for each task based on the `input\_subject' and the strength\_dict of the candidate models.\\
Step2: For the second route\_list in the `input\_route\_lists', \\
\text{\quad}- If this route\_list is the same as the first one, select the second-best model for each task based on the `input\_subject' and the strength\_dict of the candidate models.\\
\text{\quad}- Otherwise, select the best model for each task based on the `input\_subject' and the strength\_dict of the candidate models.\\
\\
\textbf{\# Output Template}\\
Repalce Variable in `\{\{\}\}'\\
Please generate the corresponding two model\_lists based on following template:
Two generated model\_lists: [\{\{selected\_model\_name1\}\}, ...], [\{\{selected\_model\_name1\}\}, ...] 
\end{tcolorbox}
\caption{Prompt template for CMS.}
\label{fig:model}
\end{figure}

\begin{figure}[ht]
\scriptsize
\centering
\begin{tcolorbox}
\textbf{\# Your task}
You are an expert in evaluating the quality of generated videos.
\\
\textbf{\# Input data}\\
`user\_input\_text': \{text\_prompt\}.\\
`user\_input\_image': \{image\_prompt\}.\\
`generated\_video': \{image\_generated\}. \\
\\
\textbf{\# Scoring pipeline.}
Step1: Based on the `user\_input\_text' and `user\_input\_image', evaluate the intrinsic quality of the `generated\_video' based on its visual aesthetics and physical plausibility.\\
Step2: Evaluate the prompt alignment of the `generated\_video' with the `user\_input\_text' and `user\_input\_image', focusing on the visual alignment and motion alignment.\\
Step3: Combine the evaluations from the previous steps to determine the overall quality of the `generated\_video', assigning a score from 1 to 20, where 1 indicates the lowest quality and 20 represents the highest.\\
\\
\textbf{\# Score}
\end{tcolorbox}
\caption{Prompt template for VQE.}
\label{fig:quality}
\end{figure}
As introduced in the main text, the video generation process of our SPAgent consists of three core modules: Decoupled Intent Recognition (DIR), Principle-guided Route Planning (PRP), and Capability-based Model Selection (CMS), and the SPAgent is equipped with the video quality evaluation module to enable its automatically tool library expansion capability. Here, we provide the prompts used in each module as shown in \cref{fig:datatype_subject} to \cref{fig:quality}, respectively. Specifically, the DIR module employs two prompts: one (\cref{fig:datatype_subject}) predicts the input data type and the subject of the video to be generated , while the other (\cref{fig:aligment}) predicts the semantic relationship between text and visual inputs, improving prediction accuracy.

\section{Dataset Curation}
\label{app:dataset}

\begin{figure}[ht]
\scriptsize
\centering
\begin{tcolorbox}
Describe the image/video concisely and accurately in a single sentence, include:\\
1. The main subject\\
2. The action or event\\
3. The location and/or time\\
Avoid excessive adjectives and focus on clarity and brevity.\\
The image/video features: \{image\_or\_video\}
\end{tcolorbox}
\caption{Prompt template for descriptive prompts.}
\label{fig:caption}
\end{figure}

\begin{figure}[ht]
\scriptsize
\centering
\begin{tcolorbox}
Given a caption, your task is to generate 3 different edits of the caption: a foreground edit, a background edit, and a style edit. Each edit should modify a specific aspect of the original caption. Identify the relevant words in the original caption and replace them with appropriate alternatives. Return the 3 edits in JSON format.
- Foreground Edit: Modify an attribute (e.g., color, shape, type) of the main object or subject in the foreground.
- Background Edit: Change an element of the background or setting described in the caption.
- Style Edit: Alter the overall style or artistic representation of the scene by adding or replacing style-related words to change the visual aesthetic.

For each edit, provide the following information in JSON.
- "type": The type of edit (foreground, background, or style)
- "source\_words": The words from the original caption being replaced (use null for style if adding new words)
- "target\_words": The new words replacing the source words (or describing the new style)
- "target\_caption": The full edited caption incorporating the changes
Ensure that each edit is distinct, diverse, and creative while maintaining plausibility and coherent within the described scene.
\end{tcolorbox}
\caption{Prompt template for editing prompts.}
\label{fig:editing}
\end{figure}

\noindent\textbf{Image and Video Input Prompts.}
After collecting the image and video, we generate one descriptive prompt and three editing prompts for image-to-video and video-to-video generation, respectively. We then use ChatGPT-4o to generate descriptions of the image and video content, as outlined in \cref{fig:caption}.
Using these descriptive prompts, we generate three editing prompts for each input: foreground, background, and style adjustments, as illustrated in \cref{fig:editing}.

\vspace{6pt}
\noindent\textbf{Output Annotation.}
In the main body of this paper, we outlined the dataset annotation process primarily from two perspectives: Intrinsic Quality and Prompt Alignment. Here, we dive deeper into the specific composition and definitions of our annotation data.
\noindent\textbf{1) Intrinsic Quality} primarily evaluates whether the generated video appears natural and coherent, encompassing two dimensions: \texttt{visual aesthetics} and \texttt{physical plausibility}. 
\noindent\textbf{2) Prompt Alignment} focuses on verifying whether the generated content and temporal changes of the output video align with the input prompt text, including two dimensions: \texttt{visual alignment} and \texttt{motion alignment}. 
Each annotator will give a final \texttt{opinion score} based on the above dimensions and their preference. Detailed definitions of the aforementioned dimensions are provided in~\cref{tab:annotation_def}. Except for the \texttt{opinion score}, which is annotated between $1$ and $20$, the other scores are between $1$ and $10$, and the higher score indicates higher preference.

\section{Case Study}
\label{app:cases}
\begin{figure*}[tp]
    \centering
    \includegraphics[width=\linewidth]{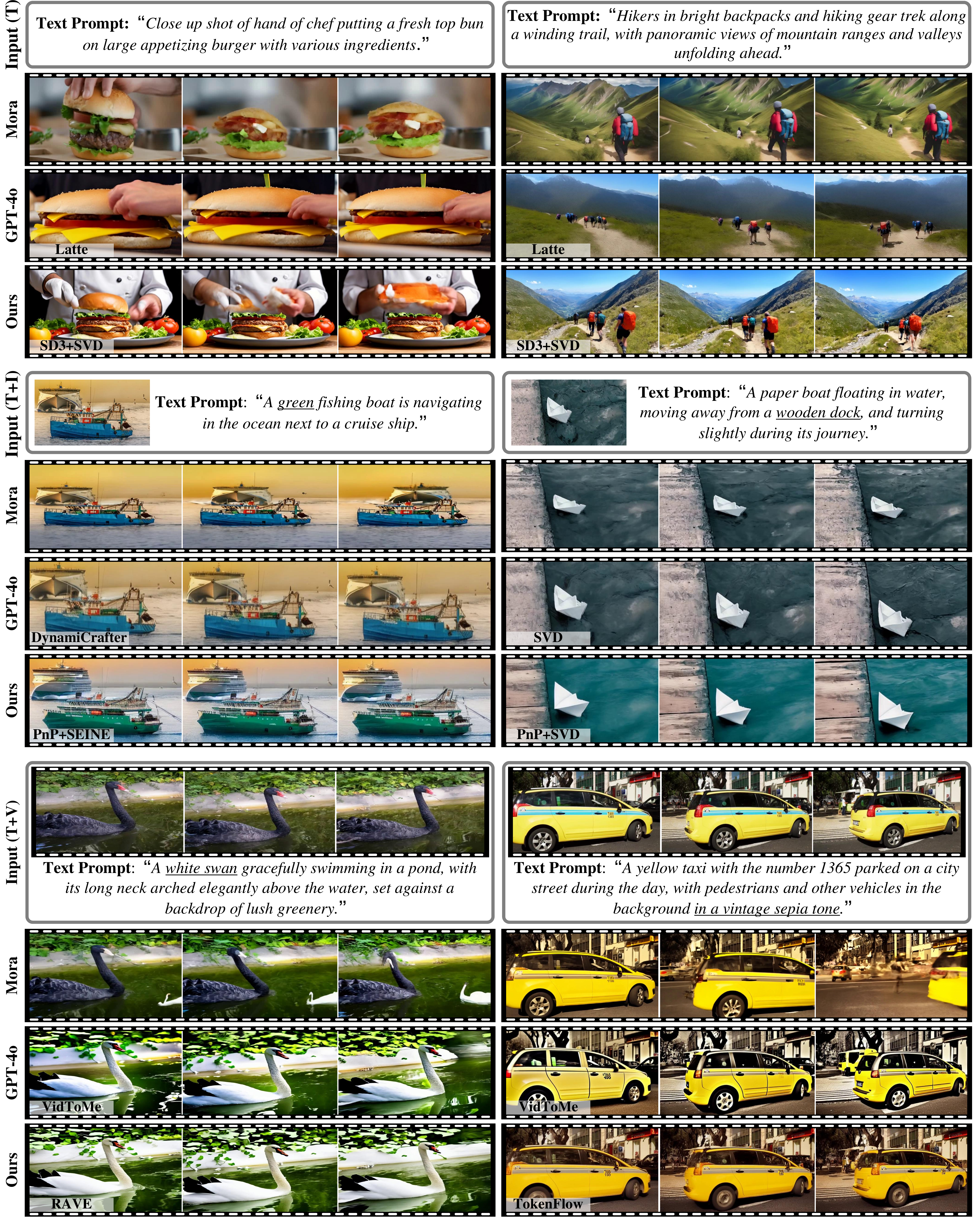}
    \caption{
         Comparison of the output videos generated by Mora~\cite{yuan2024mora}, GPT-4o~\cite{hurst2024gpt-4o} and our SPAgent given different types of user input data.
    }
    \label{fig:appendix_case_study}
\end{figure*}

\begin{figure*}[tp]
    \centering
    \includegraphics[width=\linewidth]{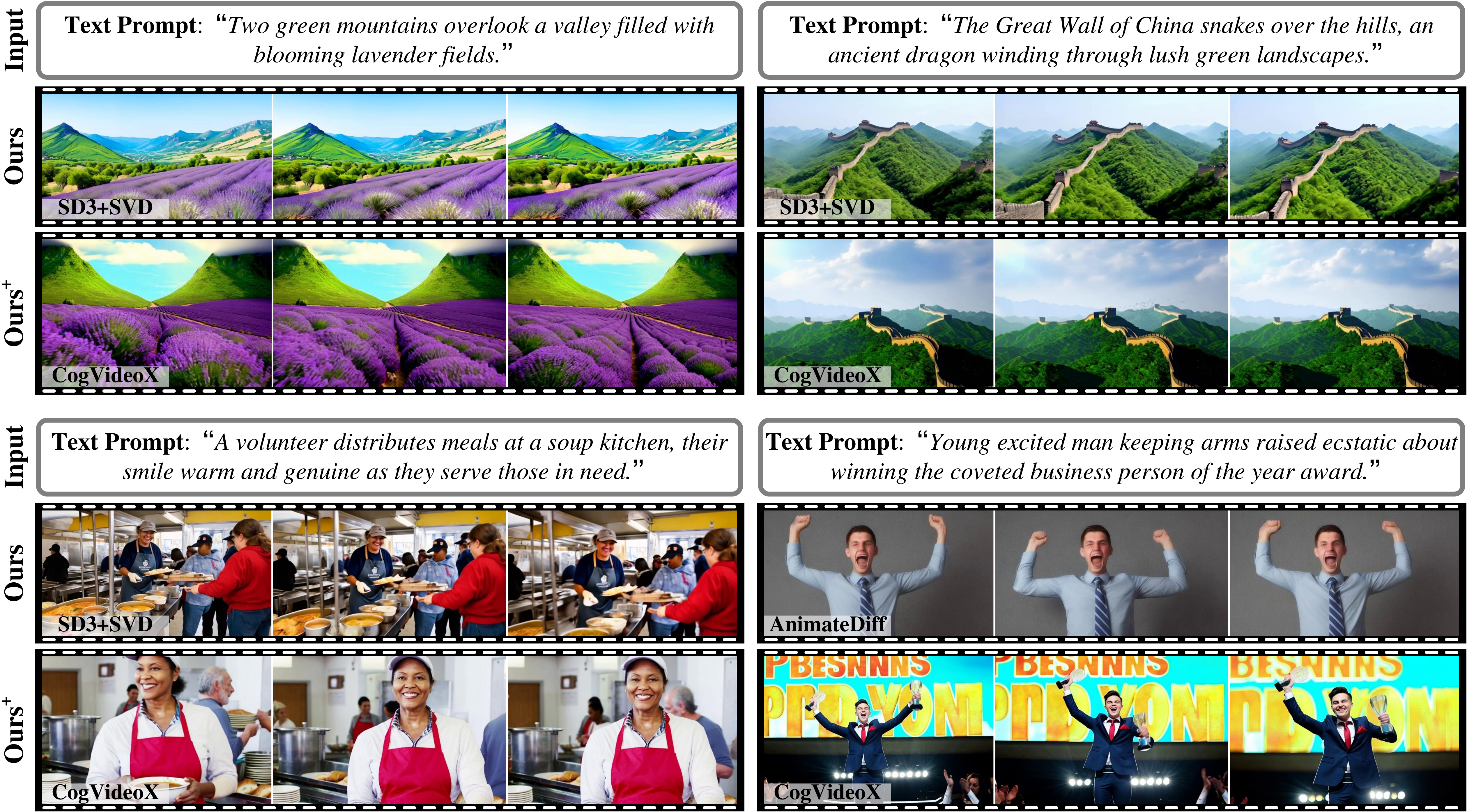}
    \caption{
         More output video samples generated by SPAgent with and without integrating CogVideoX~\cite{yang2024cogvideox} into its tool library
    }
    \label{fig:appendix_tool_expansion}
\end{figure*}

In this section, we provide more detailed case studies to demonstrate the superior performance of our method. As shown in~\cref{fig:appendix_case_study}, we illustrate three typical types of user input data: text only (T), text with image (T+I), and text with video (T+V), each corresponding to different user needs. Besides, \cref{fig:appendix_tool_expansion} gives more samples of the output video of our SPAgent after automatic tool expansion.
From the presented examples, we can draw the following conclusions: 
\textbf{1)} Our SPAgent can select more suitable model based on specific user needs, thereby generating higher-quality videos. 

\noindent\textbf{2)} SPAgent is capable of independently determining whether the user input contains editing requirements and can automatically invoke image or video editing tools accordingly. 

\noindent\textbf{3)} Through autonomous tool expansion, SPAgent can independently evaluate the capabilities of newly integrated models and subsequently invoke them for subjects where they excel (\textit{e.g.,} landscapes, people) to enhance the quality of videos in these subjects.

\begin{table*}[t]
\centering
\caption{The detailed definitions of different annotation dimensions for the output videos in our curated dataset.}
\label{tab:annotation_def}
\resizebox{\textwidth}{!}{%
\begin{tabular}{@{}ll@{}}
\toprule
Annotation & Definition \\ \midrule
\multirow{5}{*}{\texttt{Visual Aesthetics}} & \textit{\textbf{Composition}: Evaluate the meaningfulness and balance of the layout (v.s. chaotic).} \\
 & \textit{\textbf{Clarity}: Assess the sharpness of the first frame (v.s. blurriness or pixelated).} \\
 & \textit{\textbf{Contrast}: Check if the first frame has appropriate exposure and color balance.} \\
 & \textit{\textbf{Smoothness}: Evaluate the fluidity of motion between frames (v.s. flicking motion blur, or jitter).} \\
 & \textit{\textbf{Consistency}: Assess if video quality remains stable or degrades over time.} \\ \midrule
\multirow{3}{*}{\texttt{Physical Plausibility}} & \textit{\textbf{Natural Motion}: Assess if object and character movements adhere to expected physical laws.} \\
 & \textit{\textbf{Interactions}: Evaluate the realism of interactions between objects and characters.} \\
 & \textit{\textbf{Artifacts}: Note any visual anomalies or distortions.} \\ \midrule
\multirow{3}{*}{\texttt{Visual Alignment}} & \textit{\begin{tabular}[c]{@{}l@{}}\textbf{Subject Accuracy}: Verify if the video depicts the correct type, number, and appearance of subjects \\ as specified in the prompt.\end{tabular}} \\
 & \textit{\textbf{Environmental Accuracy}: Check if the setting and background match the prompt description.} \\
 & \textit{\textbf{Style Accuracy}: Assess if the overall visual style aligns with related instructions in the prompt.} \\ \midrule
\multirow{3}{*}{\texttt{Motion Alignment}} & \textit{\textbf{Action}: Evaluate if the subjects' behaviors and actions match the prompt instructions.} \\
 & \textit{\textbf{Kinetics}: Assess if the speed and direction of movements align with the prompt.} \\
 & \textit{\textbf{Camera Work}: Verify if any specified camera movements or angles are accurately represented.} \\ \midrule
\texttt{Opinion Score} & \textit{\textbf{Considering all aspects and annotator's own preference}, a subjective score for the whole video.} \\ \bottomrule
\end{tabular}%
}
\end{table*}

